\documentclass[lettersize,journal]{IEEEtran}
\usepackage{amsmath,amsfonts,amssymb}
\usepackage{algorithmic}
\usepackage{algorithm}
\usepackage{array}
\usepackage[caption=false,font=normalsize,labelfont=sf,textfont=sf]{subfig}
\usepackage{textcomp}
\usepackage{stfloats}
\usepackage{url}
\usepackage{verbatim}
\usepackage{graphicx}
\usepackage{cite}
\usepackage{booktabs}
\usepackage{tikz}
\usepackage{multirow}
\usepackage{pifont}
\usepackage{xcolor}

\usetikzlibrary{arrows,shapes,fit,scopes,calc,matrix,positioning,decorations.pathmorphing,fadings}

\tikzset{
    vertex/.style={rectangle,draw,minimum width=6em},
    edge/.style={->,> = latex'},
    cvertex/.style={circle,draw,minimum width=0.4em,inner sep=0.8},
    lcvertex/.style={circle,draw,minimum width=1em,inner sep=0.8}    
}

\pgfmathsetmacro{\tanktwo}{-1.5}
\pgfmathsetmacro{\tap}{-0.2}

\usepackage{amsthm}
\newtheorem{example}{Example}[section]

\begin{document}

\title{Probabilistic Machine Learning for Uncertainty-Aware Diagnosis of Industrial Systems}

\author{Arman Mohammadi, Mattias Krysander, Daniel Jung, Erik Frisk
\thanks{Manuscript submitted on July 3, 2025. This work is partially sponsored by ELLIIT and Sweden's innovation agency Vinnova through the project DELPHI.}
\thanks{Arman Mohammadi is with the Division of Vehicular Systems, Link\"{o}ping University, Link\"{o}ping, Sweden (email: arman.mohammadi@liu.se). He is the corresponding author.}
\thanks{Mattias Krysander is with the Division of Vehicular Systems, Link\"{o}ping University, Link\"{o}ping, Sweden (email: mattias.krysander@liu.se).}
\thanks{Daniel Jung is with the Division of Vehicular Systems, Link\"{o}ping University, Link\"{o}ping, Sweden (email: daniel.jung@liu.se).}
\thanks{Erik Frisk is with the Division of Vehicular Systems, Link\"{o}ping University, Link\"{o}ping, Sweden (email: erik.frisk@liu.se).}
}

\maketitle

\begin{abstract}
    Deep neural networks has been increasingly applied in fault diagnostics, where it uses historical data 
    to capture systems behavior, bypassing the need for high-fidelity physical models.
    However, despite their competence in prediction tasks, these models often struggle with
    the evaluation of their confidence. This matter is particularly 
    important in consistency-based diagnosis where decision logic is highly sensitive to false alarms.
    To address this challenge, this work presents a diagnostic framework that uses
    ensemble probabilistic machine learning to
    improve diagnostic characteristics of data driven consistency based diagnosis 
    by quantifying and automating the prediction uncertainty. 
    The proposed method is evaluated across several case studies using both ablation 
    and comparative analyses, showing consistent improvements across a range of diagnostic metrics.
\end{abstract}


\section{Introduction}
\label{sec:introduction}
A common strategy, known as Consistency-Based Diagnosis (CBD), frames fault detection and isolation as
tests of whether observed behaviors are consistent with expectations.
Faults are detected when observations deviate from the fault-free behavioral model,
and isolated by checking inconsistencies across fault models.
Traditionally, the developments relied on model-based approaches,
using physical insight to generate residuals that indicate faults by
comparing sensor measurements with expected outputs \cite{blanke2006diagnosis}.
While effective,
this approach requires detailed models and extensive expert knowledge,
making it time-consuming and costly \cite{pulido2019state}.
Machine learning, and specifically data-driven regression models,
has been increasingly applied in diagnostics, where it uses historical
data to capture nominal and faulty behaviors directly,
bypassing the need for high-fidelity physical models \cite{qin2012survey}.
Recent advances in deep learning have allowed data-driven regression models to
identify subtle patterns and trends, making fault detection and
isolation possible even in complex dynamic systems \cite{s23010001}.

Despite their competence in achieving high accuracy in prediction tasks, deep neural networks
often struggle in evaluation of their confidence in their estimation.
A key factor is uncertainty, which comes in two primary forms.
On one hand, there is uncertainty intrinsic to the system,
such as the inherent noise of measurement devices or the stochastic nature
of system dynamics, referred to as \textit{aleatoric uncertainty}.
This type of uncertainty persists even with existence of infinite data;
addressing it often requires additional information beyond the available features
of the deep neural network.
On the other hand,
real-world environments frequently violate the assumption that training and testing
data share the same underlying distribution.
As a result, data-driven models often face difficulties when encountering
conditions that differ from those seen during training.
This leads to \textit{epistemic uncertainty}, which reflects the model's
lack of knowledge or exposure to certain scenarios.
Unlike aleatoric uncertainty,
epistemic uncertainty can be mitigated by collecting more diverse or representative
data that better captures the operational space of the system \cite{hullermeier2021aleatoric}.

\begin{figure}
    \begin{center}
        \includegraphics[width=\columnwidth]{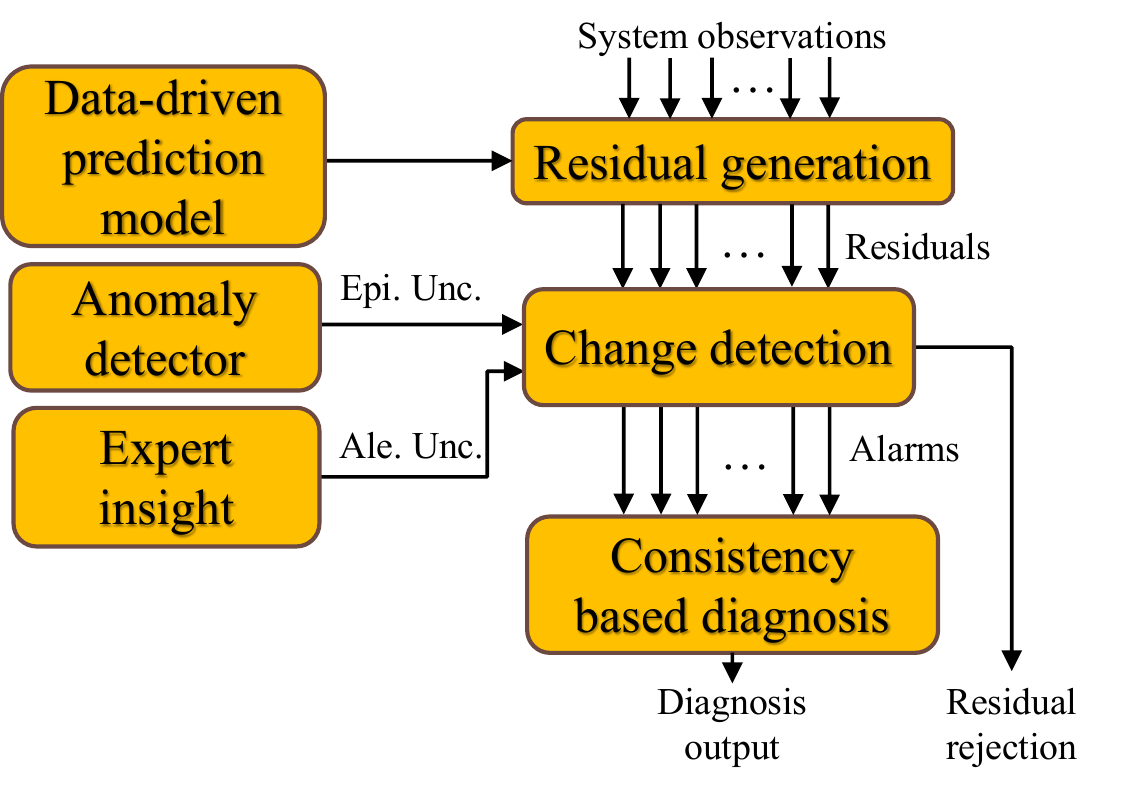}
        \caption{The proposed process by \cite{MOHAMMADI2025106283} for developments
            made for industrial applications.}
        \label{fig:process_schematic}
    \end{center}
\end{figure}

In addressing the challenges of predictive uncertainty in data-driven models,
\cite{MOHAMMADI2025106283} proposed a diagnostic framework that
addresses the issue of overly confident predictions in neural networks,
to avoid false alarms.
This is especially critical in consistency-based diagnosis, where the decision logic is highly sensitive
to false alarms.
In the absence of alarms, all fault modes remain theoretical hypotheses,
meaning the correct diagnosis still exists within the set of possible explanations.
However, when a false alarm occurs, it eliminates valid hypotheses and reinforces incorrect conclusions.
To manage uncertainty, the framework introduces separate,
measurable parameters for each uncertainty type.
Since aleatoric uncertainty represents intrinsic noise or chaos in the system output,
it should not in itself trigger an alarm but can still allow for a reliable diagnosis.
Thus, an adaptive threshold is applied to account for variations without overreacting.
In contrast, epistemic uncertainty means the model is
encountering unfamiliar conditions, so a diagnosis made under high epistemic uncertainty is unreliable.
To address this, a One-Class Support Vector Machine (SVM) is used to
detect out-of-distribution (OOD) samples and reject these residuals entirely.
Figure~\ref{fig:process_schematic} illustrates the proposed process for industrial applications.

\subsection{Problem statement and contributions}
While the existing frameworks handles predictive uncertainty in the examined case studies,
it faces certain limitations. First, the use of One-Class SVM for anomaly detection may encounter
difficulties in complex scenarios with high-dimensional data or dynamic relationships,
due to its decision boundary definitions. Additionally, the identification of key features for addressing
aleatoric uncertainty was performed empirically, which may limit
the generalization of the proposed framework.
To address these challenges, This work explores the advancements in predictive uncertainty of deep
ensemble neural networks \cite{lakshminarayanan2016decision},
to manage complex, high-dimensional data for gauging epistemic uncertainty
while providing an automated
measure for aleatoric uncertainty too.
With respect to the previous works, this paper develops a diagnostic framework that
accounts for uncertainty by combining deep ensemble neural networks with
consistency-based decision logic.
It provides practical guidance for designing and implementing the framework,
including network architecture and training agenda.
Finally, it introduces consistency-based evaluation metrics to assess
performance across both simulation and industrial case studies.

\subsection{Related research}

Early advances have focused on adapting neural networks to incorporate uncertainty
and probabilistic methods,
primarily through a Bayesian approach \cite{bernardo2009bayesian}.
This involves specifying a prior distribution
over the network's parameters and, based on training data,
computing a posterior distribution to quantify predictive uncertainty.
Since exact Bayesian inference in neural networks is computationally intractable,
various approximation techniques have been developed,
including Laplace approximation \cite{mackay1992bayesian},
Markov Chain Monte Carlo (MCMC) methods \cite{neal2012bayesian},
as well as advances in variational Bayesian methods
\cite{blundell2015weight, graves2011practical, louizos2016structured},
assumed density filtering \cite{hernandez2015probabilistic},
expectation propagation \cite{hasenclever2017distributed},
and stochastic gradient MCMC approaches like Langevin
dynamics \cite{korattikara2015bayesian} and Hamiltonian methods \cite{springenberg2016bayesian}.
In practice, Bayesian neural networks can be
more complex to implement and slower to train than non-Bayesian networks.

Gal and Ghahramani \cite{gal2016dropout} proposed using Monte Carlo dropout (MC-dropout)
to estimate predictive uncertainty by applying Dropout \cite{srivastava2014dropout}
at test time. This approach has been linked to an approximate Bayesian interpretation
of dropout, and its simplicity
has led to widespread practical use. Interestingly, dropout can also be
interpreted as a form of
ensemble model combination \cite{srivastava2014dropout},
where predictions are averaged across
an ensemble of neural networks.
This ensemble perspective may be more
applicable, particularly in cases where dropout rates are not tuned
based on the training data,
as any credible approximation to the true Bayesian posterior
should be informed by the training data.
This ensemble interpretation suggests that ensembles
could serve as an alternative approach
for estimating predictive uncertainty.

It has long been recognized that model ensembles improve predictive performance
\cite{dietterich2000ensemble}. However, it is not always clear when and why an ensemble of
neural networks can be expected to yield reliable uncertainty estimates.
Bayesian model averaging (BMA)
operates under the assumption that the true model lies within the prior's hypothesis class and
performs a soft model selection to identify the single best model \cite{minka2000bayesian}.
In contrast, ensembles combine multiple models to form a more robust predictor,
which can be beneficial when the true model does not fall within
the original hypothesis class.
Related discussions on ensemble robustness and model selection can be found in
\cite{clarke2003comparing, lakshminarayanan2016decision} and \cite{minka2000bayesian}.

\section{Background}
\label{sec:background}

In this section, the principles of consistency-based diagnosis
are summarized. Then, the design principles of
data-driven residuals using a structural model of the system are
presented. Finally, the theoretical background for
uncertainty quantification using ensemble deep neural networks with
adaptation from \cite{lindqvist2020general} is given.

\subsection{Consistency based diagnosis}

Fault diagnosis has traditionally been approached from two main directions:
control theory, commonly referred to as Fault Detection and Isolation
(FDI) \cite{gertler1990new, gertler2017fault},
and Artificial Intelligence (AI) \cite{de1987diagnosing, de1992characterizing, hamscher1992readings}.
FDI methods primarily focus on designing residual signals based on system models to
detect deviations caused by faults, while development in AI,
particularly consistency-based diagnosis, emphasize fault isolation by identifying
conflicts between expected and observed behavior.
A survey of the efforts to bridge these two communities,
including hybrid approaches and integrative frameworks, can be found in
\cite{trave2014bridging}.
In this work, we adopt a unified diagnostic framework proposed by \cite{krysander2006design},
that uses diagnostic tests,
such as residuals, designed to detect faults in specific parts of a system
while remaining insensitive to faults in other areas.

By analyzing which residuals trigger alarms, a set of potential fault candidates
can be identified, where each candidate represents a combination of faults
explaining the observed inconsistencies \cite{trave2014bridging}.
Traditionally, model-based fault isolation relies on analyzing the model structure
to determine which components are represented in each residual
\cite{pucel2009diagnosability}.
This is achieved by designing residuals with sensitivity to specific faults
while being insensitive to others,
often summarized in a fault signature matrix \cite{frisk2017toolbox}.
By integrating machine learning regression models into this
consistency-based diagnostic framework, data-driven residuals can be generated
that detect inconsistencies without requiring explicit
physical models \cite{jung2019isolation}.
Furthermore, due to the high costs and challenges associated with collecting
comprehensive fault data,
particularly in the early
stages of system development, when data from diverse fault scenarios is often
limited and unlabeled \cite{sankavaram2015incremental},
this approach trains data-driven models on only nominal system behavior.

\subsection{Residual design via structural analysis}
Structural analysis is a useful tool for methodological design and analysis of
diagnosis systems \cite{frisk2017toolbox}.
This method is particularly useful in early system design which allows for diagnosability analysis
without requiring precise parameter values \cite{blanke2006diagnosis, krysander2006design}.
A structural model can be represented as a bipartite graph
$\mathcal{M} = (\mathcal{E}, \mathcal{X},E)$ where $\mathcal{E}$ is the
set of equations, $\mathcal{X}$ is the set of variables including
all known and unknown variables and fault signals
and $E$ is a set of edges that encodes the correspondence between equations and variables \cite{6179556}.
In principle, achieving structural fault detectability depends on whether it
is possible to design a residual generator that models the part of the system affected by the fault.

By utilizing a technique known as Dulmage-Mendelsohn (DM) decomposition on the structural model,
it becomes feasible to perform various analyses,
including fault detectability and isolability assessments,
as well as the identification of redundant equation sets for residual generation \cite{krysander2007efficient}.
The DM decomposition partitions the structural model into an under-determined $\mathcal{M}^{-}$,
exactly determined $\mathcal{M}^{0}$ and an
over-determined part $\mathcal{M}^{+}$.
Different residual candidates can be constructed by identifying various
subsets of the over-determined part of the model $\mathcal{M}^{+}$ that remains
over-determined \cite{krysander2007efficient}. From a diagnosis perspective,
the minimally over-determined equation sets,
so-called minimally (structurally) over-determined (MSO) equation sets,
are of special interest since these correspond to the smallest parts of
the system that can be monitored separately \cite{krysander2007efficient}.

\begin{example}\label{ex:structural_model_example}
    To illustrate the development and analysis of a structural model for fault diagnosis,
    this example uses the three-tank system modeling case presented in \cite{frisk2017toolbox}.
    The system is described by the following set of equations:
    \begin{equation} \label{eq:dynamic}
        \begin{aligned}
             & e_1: q_1 = \frac{1}{R_{V1}} (p_1 - p_2) + f_{V1},       & \quad & e_7: y_1 = p_1,                      \\
             & e_2: q_2 = \frac{1}{R_{V2}} (p_2 - p_3) + f_{V2},       & \quad & e_8: y_2 = q_2,                      \\
             & e_3: q_3 = \frac{1}{R_{V3}} (p_3) + f_{V3},             & \quad & e_9: y_3 = q_0,                      \\
             & e_4: \dot{p}_1 = \frac{1}{C_{T1}} (q_0 - q_1) + f_{T1}, & \quad & e_{10}: \dot{p}_1 = \frac{dp_1}{dt}, \\
             & e_5: \dot{p}_2 = \frac{1}{C_{T2}} (q_1 - q_2) + f_{T2}, & \quad & e_{11}: \dot{p}_2 = \frac{dp_2}{dt}, \\
             & e_6: \dot{p}_3 = \frac{1}{C_{T3}} (q_2 - q_3) + f_{T3}, & \quad & e_{12}: \dot{p}_3 = \frac{dp_3}{dt},
        \end{aligned}
    \end{equation}
    where $f_i$ are fault signals, $y_i$ are known signals, $C_i$ and $R_i$ are known fixed parameters,
    and the remaining are unknown variables.
    Figure~\ref{fig:Structural_model_example} shows the resulting structural model,
    constructed based on the component equations in \eqref{eq:dynamic}.
    The structural model consists of 12 equations and 10 unknown variables,
    including three dynamic state variables (marked I) and their
    corresponding derivatives (marked D). Additionally, the model includes six
    fault variables and three known variables.
    Using the \textit{Fault Diagnosis Toolbox} \cite{frisk2017toolbox},
    three MSO sets that achieve maximum isolability are extracted.
    Figure~\ref{fig:FSM_FIM_example} presents the corresponding fault
    signature matrix and fault isolability matrix derived
    from the given MSO sets.
    Each row in the fault signature matrix indicates which faults influence the equations in an MSO set.
    A dot in position \((i, j)\) of the isolability matrix indicates that fault \(j\) is
    a diagnosis if fault \(i\) is the true fault.

    \begin{figure}
        \begin{center}
            \includegraphics[width=\columnwidth]{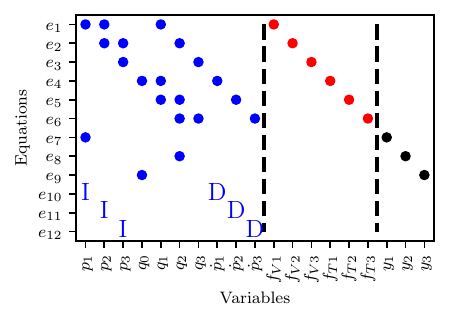}
            \caption{Structural model of the three-tank system. Equations and variables are arranged to reflect causal and dependency relationships. Dynamic variables are marked as \textbf{I} (integral states) and \textbf{D} (their time derivatives).}
            \label{fig:Structural_model_example}
        \end{center}
    \end{figure}

    \begin{figure}
        \begin{center}
            \includegraphics[width=\columnwidth]{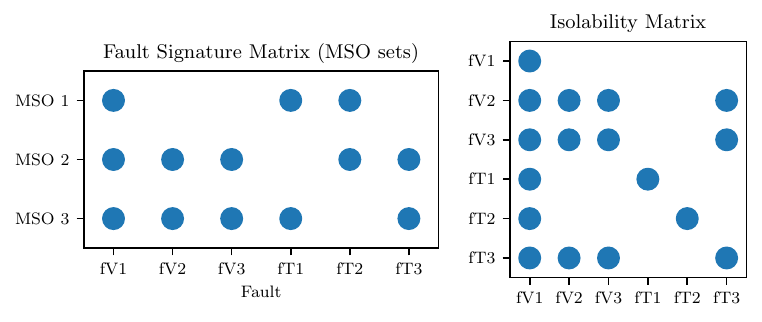}
            \caption{Fault signature matrix and isolability matrix corresponding to three MSO sets.}
            \label{fig:FSM_FIM_example}
        \end{center}
    \end{figure}

\end{example}

Several approaches exist for constructing residual generators from a given MSO set.
By removing one equation from an MSO set,
what is left has an equal number of unknown variables and equations,
creating an exactly determined set. This set can be used to determine
how to compute all unknown variables from the known ones using
a matching algorithm \cite{krysander2006design}. Then, the remaining
equation, called the residual equation, can be used to compute unknown variables.

The creation of neural network-based residuals from MSO sets involves
using neural networks to model discrete-time nonlinear representations of the states.
In this model, the network structure is designed using the computational graph attained from
the matching algorithm.
The dynamic system is formulated in a state-space model,
where unknown functions governing state transition
computations are approximated using Multilayer Perceptron (MLP) blocks.
By training residuals in an autoregressive manner on nominal system behavior,
it is possible to generate residual signals that are ideally sensitive
to specific faults, as identified through structural analysis \cite{MOHAMMADI2025106283}.

\subsection{Uncertainty quantification of neural networks using ensemble models}

Given a set of pairs of inputs and targets $\mathcal{D} = \{(x_i, y_i)\}_{i=1}^N$, a
probabilistic predictive model approximates the true
conditional probability distribution $p(y \,|\, x, \mathcal{D})$
with $q(y; f_\theta(x))$, where $q$ belongs to some family of
distributions parameterized by $f_\theta$.
In this paper, under the assumption that $q$ is a Gaussian distribution, for each input $x$,
the neural network outputs a parameter vector $z = f_\theta(x)$,
where $z = (\hat{\mu}, \hat{\sigma}^2)$ represents the predicted mean
and variance of the normal distribution.
The network parameters $\theta$ are optimized in order to maximize
the likelihood of the data with respect to $q(y; f_\theta(x))$.

The uncertainty in a model's prediction can be characterized using the estimated conditional
probability.
For a fixed value of $\theta$, the model $q(y; f_\theta(x))$ will only capture aleatoric uncertainty.
Conceptually, we can address this limitation with a Bayesian approach,
learning a posterior distribution over the model parameters $p(\theta \,|\, \mathcal{D})$
and expressing the predictive distribution for a data point $x^*$ as:
\begin{equation}
    \label{eq:baysian_illustration_distillation}
    p(y^* \,|\, x^*, \mathcal{D}) = \int \underbrace{p(y^* \,|\, x^*, \theta)}_{\text{aleatoric}} \underbrace{p(\theta \,|\, \mathcal{D})}_{\text{epistemic}} \, d\theta.
\end{equation}
More specifically, we can use this approach to define the different types of uncertainty:
\begin{equation}
    \label{eq:ensemble_distillation_logic}
    \begin{aligned}
        U_{\text{tot}} & = \mathcal{I} \left[ p(y \,|\, x, \mathcal{D}) \right],                                                    \\
        U_{\text{ale}} & = \mathbb{E}_{p(\theta \,|\, \mathcal{D})} \left[ \mathcal{I} \left[ p(y \,|\, x, \theta) \right] \right], \\
        U_{\text{epi}} & = U_{\text{tot}} - U_{\text{ale}},
    \end{aligned}
\end{equation}
where $\mathcal{I}$ is some uncertainty measure,
such as variance, entropy, or differential entropy \cite{lindqvist2020general}.
Computing the posterior distribution $p(\theta \,|\, \mathcal{D})$ is intractable when
$f_\theta$ is given by a deep neural network. A simple approach is training an ensemble of
$M$ independent models $\{ f_{\theta_m} \}_{m=1}^M$ to provide a natural
approximation of the posterior $p(\theta \,|\, \mathcal{D})$
to compute different types of uncertainty using
\eqref{eq:ensemble_distillation_logic}, an idea that has been previously explored by
\cite{lakshminarayanan2016decision,kendall2017uncertainties,malinin2019ensembledistributiondistillation}.

\section{Proposed framework}
\label{sec:proposed_framework}

First, the derivation of formulation for quantification of epistemic
and aleatoric uncertainty is shown. Then, the diagnostic framework that
integrates the ensemble uncertainty in CBD decision logic is illustrated.

\subsection{Uncertainty quantification for probabilistic regression models}
By treating an ensemble of probabilistic neural networks as an independent and uniformly
weighted mixture, we approximate the aggregated predictive distribution by matching its first and second moments.
The ensemble estimate of the predictive mean $\hat{\mu}_{\ast}$ and variance $\hat{\sigma}_{\ast}^2$
are calculated as (See Appendix~\ref{app:ensemble_moments}):
\begin{subequations}
    \label{eq:ensemble_mean_variance}
    \begin{align}
        \hat{\mu}_{\ast}      & = \frac{1}{M} \sum_{m=1}^{M} \hat{\mu}_m,     \\
        \hat{\sigma}_{\ast}^2 & = \frac{1}{M} \sum_{m=1}^{M} \hat{\sigma}_m^2
        + \frac{1}{M} \sum_{m=1}^{M} (\hat{\mu}_m - \hat{\mu}_{\ast})^2
    \end{align}
\end{subequations}
where \(\hat{\mu}_m\) and \(\hat{\sigma}_m^2\) denote the predicted mean and variance,
respectively, from the \(m\)-th model in the ensemble for a given input.
Considering the variance representation of uncertainty,
each model individually estimates the inherent variability of the output by predicting a variance $\hat{\sigma}_m^2$.
Since the models are treated as independent samples from the posterior over parameters,
the overall aleatoric uncertainty is approximated by the average of the predicted variances across all ensemble members.
This corresponds to the Monte Carlo approximation of the expected predictive variance over the posterior distribution,
as defined in \eqref{eq:ensemble_distillation_logic}.
The total predictive uncertainty is represented by the variance of the aggregated
predictive distribution $\hat{\sigma}_{\ast}^2$.
Thus, the aleatoric and epistemic uncertainties can be approximated as:
\begin{equation}
    \label{eq:uncertainty_measures}
    U_{\text{ale}} \approx \frac{1}{M} \sum_{m=1}^{M} \hat{\sigma}_m^2, \quad
    U_{\text{epi}} \approx \frac{1}{M} \sum_{m=1}^{M} (\hat{\mu}_m - \hat{\mu}_{\ast})^2
\end{equation}

\begin{example}\label{ex:ensemble_example}
    To illustrate the intuition behind this approach, consider a dataset where the training samples are drawn from the function
    $y = x^3 + \xi(x)$, where $\xi(x)$ represents noise as a function of the input $x$,
    simulating aleatoric uncertainty. The input $x$ for the training data is sampled within the range
    $x \in [-2, 2]$, whereas the test set extends beyond this range, with $x \in [-3, 3]$. This setup
    introduces epistemic uncertainty in the regions $x \in [-3, -2]$ and $x \in [2, 3]$, as these input
    values lie outside the training distribution. Figure~\ref{fig:Ensembled_example_uncertainty} illustrates this setup using
    \eqref{eq:ensemble_mean_variance} and \eqref{eq:uncertainty_measures} on 10 stacks
    of probabilistic predictive models of neural networks (referred to as PNNs throughout this paper).
    As it can be seen, aleatoric uncertainty is illustrated by the variance in
    the prediction within the training range,
    where noise $\xi(x)$ affects the data.
    Epistemic uncertainty exists in the extrapolated
    regions where the model's confidence decreases due to a lack of training data.
    \begin{figure}
        \begin{center}
            \includegraphics[width=\columnwidth]{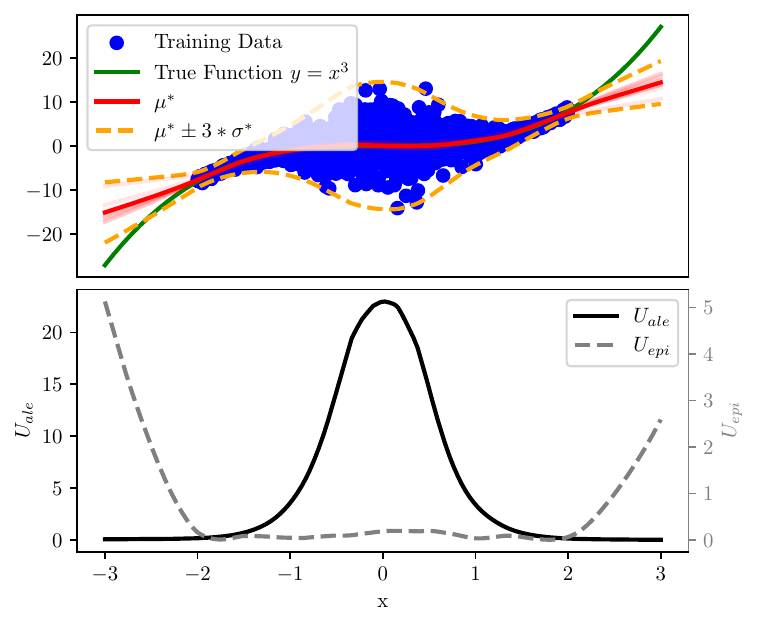}
            \caption{Illustration of the training dataset, the true function
                $y = x^3$, ensemble predictions
                $\hat{\mu}_{\ast}$ and $\hat{\sigma}_{\ast}$ of 10 stacks of neural network, and
                the corresponding aleatoric and epistemic uncertainty.
            }
            \label{fig:Ensembled_example_uncertainty}
        \end{center}
    \end{figure}
\end{example}

\subsection{Uncertainty-aware diagnostic framework}

\begin{figure}
    \begin{center}
        \includegraphics[width=\columnwidth]{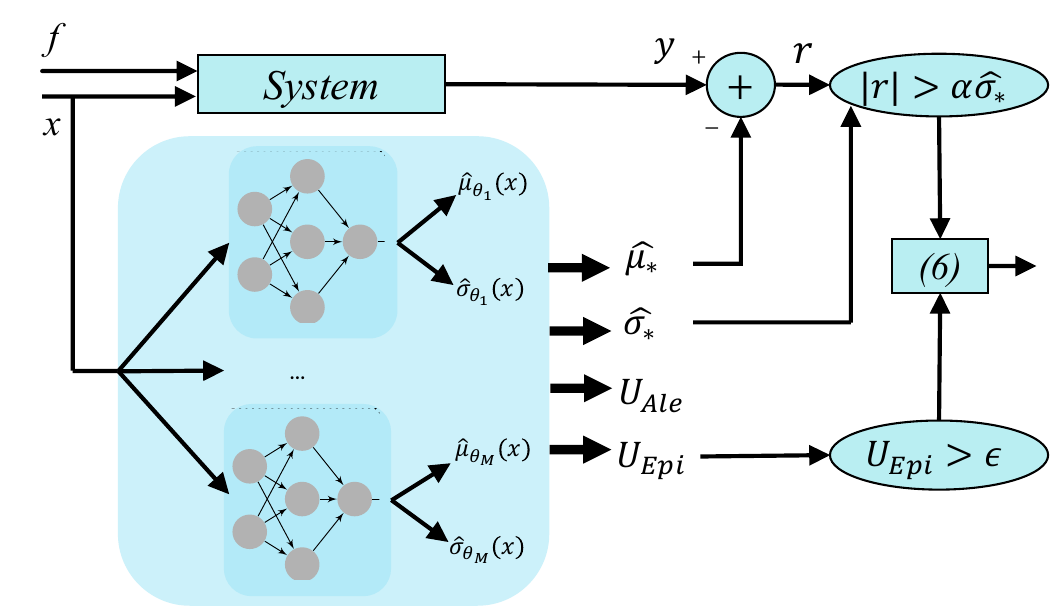}
        \caption{Overview of the proposed uncertainty-aware diagnostic framework.
            Diagnostic decisions are made based on the logic defined in \eqref{eq:decision_logic}.}
        \label{fig:Ensembled_diagnosis_framework}
    \end{center}
\end{figure}

The proposed framework builds upon the structural analysis-based
residual generation discussed earlier and an ensemble of probabilistic neural networks (PNNs) for regression tasks.
The foundation of the approach lies in maintaining the MSO-based causal structure for each residual,
with the expectation that each trained residual will exhibit the fault
sensitivity pattern assumed in the structural analysis.
Figure~\ref{fig:Ensembled_diagnosis_framework} provides a schematic view of the uncertainty-aware diagnostic framework.
Each residual
$r = y - \hat{\mu}_{\ast} \sim \mathcal{N}(0, \hat{\sigma}^2_\ast)$,
is modeled using an ensemble PNN,
which also quantifies aleatoric and epistemic uncertainty.
The decision-making process is formalized as:
\begin{equation}\label{eq:decision_logic}
    \begin{aligned}
         & U_{\text{epi}} > \epsilon                            & \text{out of range}   \\
         & |r| \leq J \quad\&\quad U_{\text{epi}} \leq \epsilon & \text{no conclusion}  \\
         & |r| > J  \quad\&\quad U_{\text{epi}} \leq \epsilon   & \text{fault detected} \\
    \end{aligned}
\end{equation}
Since uncertainty approximates the variance of the prediction, under the assumption of a Gaussian normal distribution with zero mean for the residual,
we formulate the threshold design as a two-sided statistical detection problem.
For a desired false alarm rate $P_{\text{fa}}$, the threshold is
given by $J = \hat{\sigma}_{\ast} \Phi^{-1}(1 - P_{\text{fa}}/2)$,
where $\Phi^{-1}$ is the inverse cumulative distribution function of the standard normal distribution.
We define a threshold parameter $\alpha = \Phi^{-1}(1 - P_{\text{fa}}/2)$ such
that $J = \alpha \hat{\sigma}_{\ast}$.
For a 1\% false alarm rate across the training set, this yields $\alpha = \Phi^{-1}(0.995) \approx 2.576$.
The parameter \( \epsilon \) determines when an out-of-distribution warning should be issued.
By normalizing \( U_{\text{epi}} \) with respect to its maximum value, excluding the top 1\% of anomalies in the training data,
we set \( \epsilon = 1 \).

\section{Evaluation and results}
\label{sec:Evaluation_and_results}

This section presents the evaluation of the proposed diagnostic approach.
It begins with a description of the datasets and case studies,
followed by implementation and training agenda of the ensemble probabilistic neural network.
The diagnostic performance is then analyzed through a set of defined evaluation metrics.
Finally, an ablation and comparative study is
conducted to further assess the contribution of each component within the diagnostic
framework and compare its performance against existing methods.

\subsection{Data sets and case study description}

Several diagnostic benchmarks are included in this study.
The requirements for using them are the availability of a structural model,
training data representing nominal system conditions, and test data from various fault scenarios.

\subsubsection{Simulation of a two-tank system}
Data and faults are generated in Simulink.
The benchmark consists of a classic two-tank water system,
where water is pumped into the upper tank and flows into the lower tank through
interconnected pipelines. Fault candidates include actuator faults in the pump,
sensor faults in level and flow measurements, leakages at various points,
and partial obstructions in the connecting pipes.
This benchmark is recognized and has been used extensively in the fault diagnosis literature,
making it a relevant baseline for comparison. Its simulation-based nature allows
controlled generation of data with desirable properties for research investigations \cite{MOHAMMADI20232909}.
A detailed schematic and fault descriptions are provided in Appendix~\ref{app:two_tank}.

\subsubsection{Aftertreatment in a heavy truck}
This industrial case study is the Selective Catalytic Reduction (SCR)
system used to reduce NOx emissions in diesel engines. The focus is on the
subsystem responsible for dosing urea into the exhaust stream.
Data is collected in a workshop environment under various drive cycles and fault conditions,
primarily targeting clogging faults. The setup includes signals from
pressure sensors and the Electronic Control Unit (ECU).
Data from different fault scenarios
are generated by replacing faulty components in the experimental setup.
The collected fault scenarios considered in this
study are motivated by previous workshop experiences.
This benchmark is included due to its experimental nature,
serving as a realistic setting to evaluate the diagnostic algorithms.
Since the dataset is pre-logged, it poses practical challenges typical of real-world applications,
such as uncontrolled conditions and limited ability to influence fault scenarios \cite{MOHAMMADI2025106283}.
Details including the system schematic, fault types and data collection methodology
are provided in Appendix~\ref{app:AT}.

\subsubsection{Air-path of an automotive gasoline engine}
This case study focuses on the air path of a
turbocharged gasoline engine, a critical subsystem for fuel injection,
emissions control, catalyst protection, and efficiency optimization.
Due to strong interactions via the exhaust turbo and intake compressor,
faults in almost any part of the air path affect multiple measured variables.
The benchmark dataset, introduced in \cite{jung2024benchmark},
was collected from an engine test bench and includes data from
nominal operations and various fault scenarios, such as sensor faults and intake manifold leakage.
This dataset is publicly available and has been established as a benchmark,
making it suitable for reproducible evaluation.
Moreover, the system's complexity and high-dimensional input space,
along with its erratic behavior under fault conditions,
pose a challenge to diagnostic algorithms.
Detailed descriptions, system schematics and fault scenarios
are provided in Appendix~\ref{app:engine_airpath}.

\subsection{Training and implementation details of the ensemble probabilistic neural network}
The core model architecture employed in this work is a recurrent neural network
designed for probabilistic sequence modeling. The model consists of a single-layer
Long Short-Term Memory (LSTM) network followed by two separate fully connected layers:
one for estimating the mean and another for the standard deviation of the predictive distribution.
Three different data sets mentioned in previous section were used for training, validation and testing,
each representative of a different diagnostic benchmark.
The nominal and fault free operation is used during training and validation,
while fault data and other set of nominal data is used for testing of the diagnostic performance.
All implementations are performed in Python using the PyTorch library.
The optimizer used is Adam.
Batch size, learning rate, weight decay rate, hidden dimensionality of network and total number of epochs
are tuned independently for each experiment.

\subsubsection{Scheduling of the training objective}
The overall training procedure is summarized in Algorithm~\ref{alg:training}.
In this work, the training objective is scheduled to change from Mean Squared Error
(MSE) to Negative Log-Likelihood (NLL) under a Gaussian distribution assumption.
This scheduling helps to warm up the model for the regression task before transitioning
to likelihood-based training. Additionally, during the warm-up phase the prediction horizon is
increased gradually, to allow the model to have a stable autoregressive prediction for full
sequence length provided in the training set.
Let $H$ denote the full prediction horizon, $H_{\text{init}}$ the initial horizon, and $\Delta H$ the step size by which the horizon is increased during training.
The model parameters are denoted as $\theta_m = \theta_m^{(\mu)} \cup \theta_m^{(\sigma)}$, where $\theta_m^{(\mu)}$ includes the parameters of the LSTM and the feedforward layer responsible for predicting the mean of the output distribution $\hat{\mu}$. The remaining parameters, $\theta_m^{(\sigma)}$, correspond to the feedforward layer that models the standard deviation of the distribution $\hat{\sigma}$.
The warm-up phase lasts for $\tau_{\text{w}}$ epochs, during which the model is trained with
$\mathcal{L}_{\mathrm{MSE}} = \frac{1}{n} \sum_{i=1}^n (y_i - \hat{\mu}_i)^2$
in which $\theta_m^{(\sigma)}$ are frozen.
After the warm-up phase, the training objective is switched to,
\[
    \mathcal{L}_{\mathrm{NLL}} = \frac{1}{n} \sum_{i=1}^n \left( \frac{(y_i - \hat{\mu}_i)^2}{2\hat{\sigma}_i^2} + \frac{{}\log \hat{\sigma}_i^2}{2} \right) + \text{const}.
\]
where $\theta_m^{(\mu)}$ are frozen and $\theta_m^{(\sigma)}$ are trained for $\tau$ epochs.
\begin{algorithm}
    \caption{Two-phase training with horizon increase}
    \label{alg:training}
    \begin{algorithmic}[1]
        % \REQUIRE Parameters $\{\theta_m\}_{m=1}^M$, epochs $\tau_{w}, \tau$, horizons $H_{\text{init}}, H$, step $\Delta H$, steps $n$
        \ENSURE Trained parameters $\{\theta_m\}_{m=1}^{M}$
        \STATE Initialize each $\theta_m$
        \STATE $H_{\text{curr}} \gets H_{\text{init}}$
        % \COMMENT{All training steps below are applied independently for each $m = 1,\dots,M$}

        \FOR{$e = 1$ to $\tau_{w} / n$}
        % \STATE Freeze $\theta_m^{(\sigma)}$
        \STATE Train each $\theta_m^{(\mu)}$ ($\mathcal{L}_{\mathrm{MSE}}$) over horizon $H_{\text{curr}}$ \\
        \STATE $H_{\text{curr}} \gets \min(H_{\text{curr}} + \Delta H, H)$
        \ENDFOR

        \FOR{$e = 1$ to $\tau$}
        % \STATE Freeze $\theta_m^{(\mu)}$
        \STATE Train each $\theta_m^{(\sigma)}$ ($\mathcal{L}_{\mathrm{NLL}}$) over full horizon $H$ \\
        % \hfill (freeze $\theta_m^\mu$)
        \ENDFOR

        \RETURN $\{\theta_m\}_{m=1}^{M}$
    \end{algorithmic}
\end{algorithm}

\subsection{Results}

The results presented in this section are derived based on the following stages.
First, for each case study, a structural analysis is conducted to identify a
set of candidate residual configurations.
Subsequently, each network is trained using
Algorithm~\ref{alg:training} on a nominal dataset, which represents the corresponding
subsystem under normal operating conditions, free of faults.
The results shown in this section are obtained by evaluating the trained models on new
datasets containing both nominal and faulty system behavior.
Due to the nature of consistency-based diagnosis, which relies on residuals in an autoregressive
fashion, all evaluations are performed on the full sequence of data.
% This ensures that the temporal dependencies inherent in the diagnostic
% process are accounted for during model assessment.

\begin{figure}
    \begin{center}
        \includegraphics[width=\columnwidth]{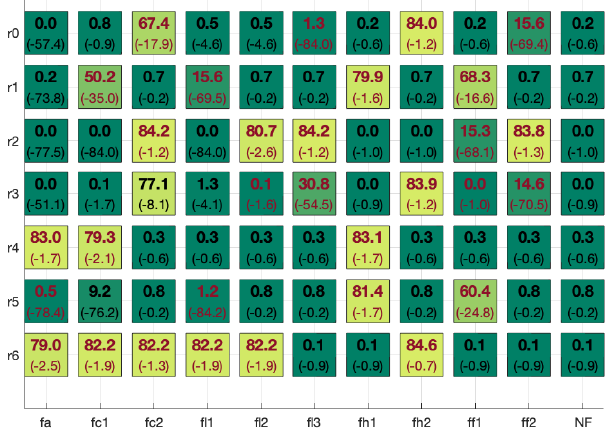}
        \caption{Probability of alarm for each residual with respect to
            different data sets used in the analysis for the simulation study. Black numbers indicate
            decoupled and red numbers indicate sensitive faults according
            to the decision structure.
            The number in parentheses corresponds to the change compared to the traditional alarm approach.}
        \label{fig:two_tank_sensitivity}
    \end{center}
\end{figure}

\begin{example}\label{ex:illustrative_example}
    Figure~\ref{fig:illustrative_example_results} illustrates a representative output of the
    diagnostic framework using one example residual from the aftertreatment system.
    The figure consists of a $3\times3$ grid, where each column corresponds
    to a different operating mode. The left column represents nominal behavior,
    the middle column a fault that is sensitive to the selected residual,
    and the right column a fault that is structurally decoupled from the residual.
    The first row shows the residual signals and their associated adaptive thresholds.
    The second row presents the epistemic uncertainty estimated by the model,
    with a black dashed line as the OOD detection threshold.
    The third row visualizes the resulting decision according to \eqref{eq:decision_logic}.
    The left column shows minimal anomalies and false alarms,
    as the nominal data aligns well with the training distribution.
    The middle column, a sensitive fault, shows both alarms and anomalies,
    since the fault causes a shift outside the valid range. The right column,
    a decoupled fault, mainly triggers OOD rather than alarms.

    \begin{figure}
        \begin{center}
            \includegraphics[width=\columnwidth]{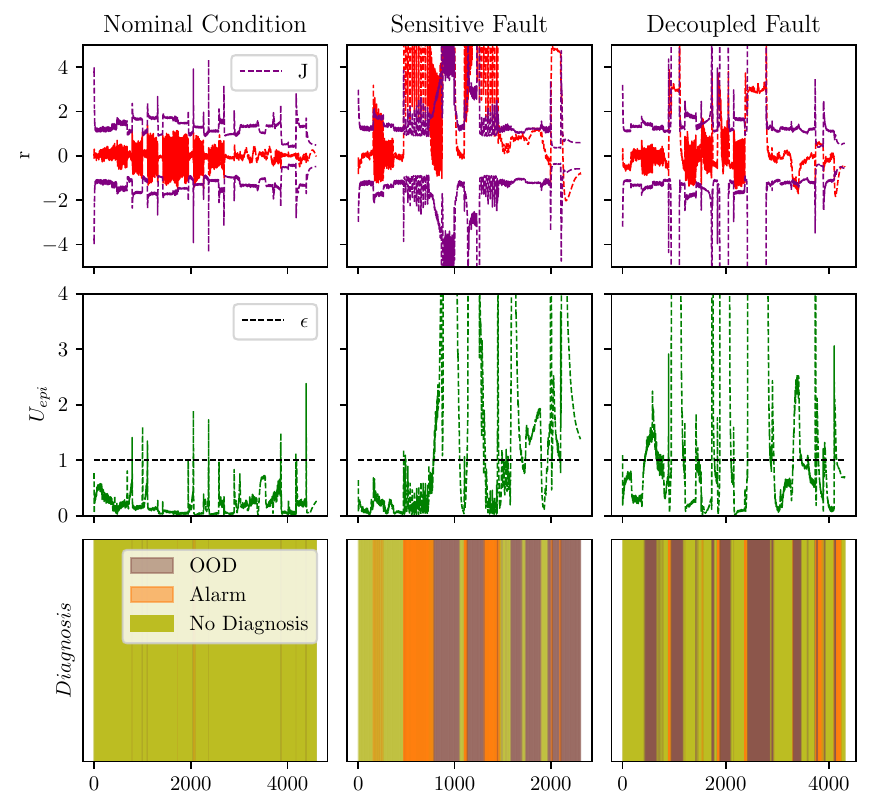}
            \caption{Illustration of an example of a residual signals, adaptive thresholds, and epistemic uncertainty across
                different fault scenarios from the aftertreatment system.
            }
            \label{fig:illustrative_example_results}
        \end{center}
    \end{figure}
\end{example}
To get an overall perspective, two performance matrices are computed.
The first summarizes the residual-level alarm behavior (corresponding to the detection of
faults according to fault signature matrix) and
the second matrix captures the diagnostic-level performance of the
overall system after applying the consistency-based decision logic considering all the residuals
(corresponding to the isolability matrix).
Figure~\ref{fig:two_tank_sensitivity} represents the residual-level alarm behavior
applied to the two-tank system.
The number in position \((i, j)\) is the probability \( s_{ij} \) in percent that
fault \( f_j \) enriches the diagnosis in \( r_i \), e.g. in case of using the framework decision
logic,
\[
    s_{ij} = P(|r_i| > J_i  \quad\&\quad U_{\text{epi,i}}\leq \epsilon \mid f_j \text{ true fault}).
\]
The fault sensitivity of each residual,
represented by the fault signature matrix, is encoded in the color of the numbers.
Black numbers indicate faults that are expected to be decoupled,
meaning the alarm percentage should be close to zero,
while red numbers indicate faults that should be detected by the residuals.
In order to provide a sense of comparison,
each cell also includes the difference in probability (in parentheses) when using the proposed
decision framework, compared to traditional alarm generation based on a prediction model with a
fixed threshold corresponding to a 1\% false alarm rate across the training set,
without any anomaly detection mechanism.
As shown in Figure~\ref{fig:two_tank_sensitivity}, a clear reduction in
alarm rates is observed across both sensitive and non-sensitive faults.
% following the application of the proposed framework.
It is important to note that the primary objective of the framework
is to reduce false alarms while preserving the true detection capability
of each residual for its associated fault cases.
This behavior is observed across several instances for example,
false alarms in faults $f_a$, $f_{c1}$ and $f_{l1}$ in
residuals $r_0$, $r_1$, $r_2$, $r_3$ and $r_5$ are largely reduced.
The only notable instance of low performance is observed in the case of
fault $f_{c2}$ in residual $r_3$, where the method fails
to suppress false alarms. This is due to the neural network not capturing
the true dynamics between the signals, instead learning statistical correlations
that do not reflect the actual system behavior.
This leads to persistent false alarms, an aspect that is discussed
in detail in Section~\ref{sec:Discussion}.

\begin{figure}
    \begin{center}
        \includegraphics[width=\columnwidth]{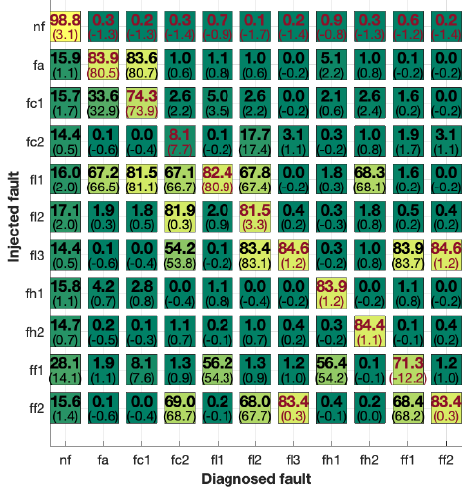}
        \caption{The fault isolation performance matrix for the simulation study.
            Black numbers indicate
            decoupled and red numbers indicate sensitive faults according
            to the isolability matrix.
            The number in parentheses corresponds to the change compared to
            the traditional alarm apprach.}
        \label{fig:two_tank_isolation_performance}
    \end{center}
\end{figure}

In order to quantify diagnosis performance,
all residual generators have been run on all data, minimal
diagnoses have been computed for each sample.
The diagnosis results for the two tank system are summarized in a fault isolation performance
matrix shown in Figure~\ref{fig:two_tank_isolation_performance}.
The number in position \((i, j)\) is the probability \( p_{ij} \) in percent that
fault \( f_j \) is a diagnosis given that the true fault is \( f_i \), i.e.,
\[
    p_{ij} = P(f_j \text{ diagnosis} \mid f_i \text{ true fault}).
\]
The evaluation results in Figure~\ref{fig:two_tank_isolation_performance} can
then be considered as the performance of the diagnosis system design using CBD decision logic.
The values in parentheses indicate the difference in performance between the
proposed framework and the traditional approach based on fixed-threshold
alarms without anomaly detection.
The structural isolation property of the system, as derived from the structural model,
is encoded using colors with red indicates fault modes that are structurally sensitive and black
for isolated faults.
Note that consistency-based diagnosis is not a classification approach
computing exactly one diagnosis. In consistency-based diagnosis, modes are
rejected if there is enough evidence to do so.
This means that the row sum in the matrix can be greater than 1, even though the ideal scenario
is that the matrix is diagonal.
The results show a clear improvement in the detection of
faults $f_a$, $f_{c1}$, and $f_{l1}$, due to a reduction in false alarms.

A similar evaluation is performed on the other two case studies, and the corresponding
result tables can be found in Appendix~\ref{app:AT} and~\ref{app:engine_airpath}.
Given the extensive number of evaluations and results presented in this study,
a set of comparative metrics is introduced to provide a systematic assessment
of diagnostic model performance.
% After that, a comparative and ablation study is conducted.

\subsection{Comparison metrics}
Selecting an appropriate diagnostic metric extends beyond simply
evaluating loss function values.
Although achieving a low training and validation loss provides an initial indication of
effective model training, diagnostic performance must also encompass
reliable fault detection and isolation.
In the consistency-based framework adopted here, the model's performance is
therefore assessed by examining its performance along several diagnostic metrics.

Consider fault signature matrix \( T_{ij} \) which corresponds to the
sensitivity for each residual \( i \) to fault data label \( j \)
(e.g. Figure~\ref{fig:FSM_FIM_example}).
Let \( N_0 \) denote the number of pairs \((i, j)\) for which \( T_{ij} = 0 \)
and \( N_1 \) the number of pairs \((i, j)\) for which \( T_{ij} = 1 \).
The average false alarm rate is denoted as
\begin{equation} \label{eq:alarm_rate}
    S_{\text{FA}} = \frac{1}{N_0} \sum_{i,j : T_{ij} = 0} s_{ij}
\end{equation}
and the average missed detection denoted rate is denoted as
\begin{equation} \label{eq:missed_detection_rate}
    S_{\text{MD}} = 1 - \frac{1}{N_1} \sum_{i,j : T_{ij} = 1} s_{ij}.
\end{equation}
These metrics are evaluating the deviation of residuals in behaving corresponding to their
decision table.
The next metrics are inspired by CBD interpretation of alarms
proposed by \cite{FRISK2018139}.
Let \( D \) denote the set of diagnoses, \( n_f \) the number of faults,
\( \tilde{F} \) the set of fault modes without the no-fault mode,
and \( I_{ij} \) the value in the structural isolability matrix in
position \( (i,j) \) (e.g, Figure~\ref{fig:FSM_FIM_example}).
The diagnosis performance measures considered here are the false alarm probability
\begin{equation} \label{eq:false_alarm}
    p_{\text{FA}} = 1 - P(NF \in D | NF),
\end{equation}
the mean missed detection probability
\begin{equation} \label{eq:missed_detection}
    p_{\text{MD}} = \frac{1}{n_f} \sum_{f_i \in \tilde{F}} P(NF \in D | f_i),
\end{equation}
and aggregated detection error
\begin{equation} \label{eq:isolation_error}
    p_{\text{D}} = \frac{1}{n_f^2} \sum_{f_i \in \tilde{F}} P(NF \not\in D | f_i)\sum_{f_j : I_{ij} = 1} \left| P(f_j \in D | f_i) - I_{ij} \right|
\end{equation}
that ideally should be 0.
While additional diagnostic metrics such as time to detection and isolation
may offer further insights,
these are not emphasized as primary comparison metrics,
given their heavy dependence on the particular detection scheme employed
in post-alarm processing stages.

\subsection{Ablation and comparative study}

To gain insight into the contributions of different components within our work,
we have conducted an ablation study. The objective of this study is to
assess the impact of key elements on model performance by
removing specific mechanisms and analyzing the resulting behavior.
To ensure fairness we excluded any usage of evaluation data,
including any data containing fault modes, from the design and tuning process.
Some metrics, such as missed detection, can only be meaningfully compared when other metrics,
like false alarms, are held constant. However, achieving such consistency
requires a complex and multidimensional thresholding and evaluation process that would inherently
involve evaluation data. To prevent this and remain true to real-world diagnostic design
principles of this study, we focused on designing models using only the nominal training data.
Each ablation experiment was performed independently,
with the modified models trained and evaluated under identical
conditions. Two components were considered here:

\begin{itemize}

    \item \textit{Out of distribution}: Epistemic uncertainty,
          which captures model uncertainty due to limited data, is estimated using
          an ensemble of PNNs and used as a measure of OOD. To assess its significance,
          we followed the original CBD logic using only alarms.

    \item \textit{Adaptive thresholding}: $J$ is modeled
          through the predicted variance in each PNN. To examine its impact,
          we replaced the adaptive thresholding with a
          fixed threshold selected to yield a 1\%
          false alarm rate on the training set.

\end{itemize}

As a point of comparison with prior work in the literature \cite{MOHAMMADI2025106283},
we replaced the OOD indicator with a SVM anomaly detector.

\begin{itemize}

    \item \textit{SVM-based anomaly detection (comparative study)}:
          The One-Class SVM model was tuned to achieve a 1\% anomaly rate on the model input of the
          training set, corresponding to the anomaly level captured by the epistemic uncertainty measure.
          % \subsection{Comparative results}
\end{itemize}

Table~\ref{tab:ablation_results} presents the results of the ablation study.
Removing both components
influences the most critical metrics, the false alarm rate ($S_{\text{FA}}$)
and the probability of false alarm ($p_{\text{FA}}$). Eventhough, this comes at the cost
of an increased missed detection ($S_{\text{MD}}$ and $p_{\text{MD}}$), the
aggregated detection error ($p_{\text{D}}$) has its lowest value in the
presence of both components in decision logic.
The same pattern is observed for removal of each individual component accross all case studies.

The comparison study reveals that the performance of ensemble probabilistic regression models
and SVM-based anomaly detection can vary across different application domains.
The ensemble method outperforms in the engine datasets,
while in the two-tank and aftertreatment
system datasets, the SVM demonstrates better performance.
A consideration when comparing ensemble models with alternative anomaly
detection methods, such as SVMs, lies
in how these models handle deviations in the input space.
One-Class SVM enforces boundaries around the training distribution,
flagging any deviation as an anomaly.
In contrast, ensemble method estimates uncertainty based only on learned patterns.
If a feature was not influential during training,
deviations in that feature will not necessarily trigger high uncertainty,
even if the feature shift indicates an anomaly.
This can lead to blind spots where anomalies go undetected.
SVM, on the other hand, explicitly models the entire feature
distribution and detects anomalies regardless of whether a feature was
important during training.
However, a notable limitation of SVM-based approaches is their susceptibility to the
curse of dimensionality. As the input dimensionality increases,
the data becomes sparser, distance metrics become less informative,
and decision boundaries become less reliable, which can degrade
performance in high-dimensional feature spaces \cite{zheng2022anomaly}.

\begin{table*}
    \centering
    \caption{Ablation study results across different scenarios. A \checkmark indicates the presence of the component, while an \textcolor{red}{\ding{55}} indicates its removal.}
    \label{tab:ablation_results}
    \begin{tabular}{c c c c|c c|c c c}
        \toprule
        System & Scenario & \textit{OOD}               & \textit{Adaptive J}        & \textbf{$S_{\text{FA}}(\%)$} & \textbf{$S_{\text{MD}}(\%)$} & \textbf{$p_{\text{FA}}(\%)$} & \textbf{$p_{\text{MD}}(\%)$} & \textbf{$p_{\text{D}}(\%)$} \\
        \midrule
        \multirow{3}{*}{Two-tank}
               & T1       & \checkmark                 & \checkmark                 & \textbf{2.224}               & 43.828                       & \textbf{1.206}               & 16.753                       & \textbf{1.132}              \\
               & T2       & \checkmark                 & \textcolor{red}{\ding{55}} & 2.790                        & 43.245                       & 4.221                        & 15.718                       & 1.365                       \\
               & T3       & \textcolor{red}{\ding{55}} & \checkmark                 & 9.112                        & 23.047                       & 1.206                        & 15.075                       & 3.498                       \\
               & T4       & \textcolor{red}{\ding{55}} & \textcolor{red}{\ding{55}} & 14.103                       & \textbf{21.138}              & 4.321                        & \textbf{14.241}              & 4.230                       \\
        % \hdashline[0.5pt/1.5pt]                                                                                                                                                                                                                                       \\[-2ex]
               & T5       & \textcolor{black}{SVM}     & \checkmark                 & \textbf{1.837}               & 45.022                       & 1.206                        & 17.708                       & \textbf{0.938}              \\
        \midrule
        \multirow{3}{*}{Aftertreatment}
        % Variance representation of epistemic
               & AT1      & \checkmark                 & \checkmark                 & \textbf{3.950}               & 72.870                       & \textbf{3.652}               & 50.886                       & \textbf{2.895}              \\
               & AT2      & \checkmark                 & \textcolor{red}{\ding{55}} & 8.681                        & 66.083                       & 11.789                       & 41.899                       & 5.051                       \\
               & AT3      & \textcolor{red}{\ding{55}} & \checkmark                 & 4.657                        & 64.002                       & 3.791                        & 47.890                       & 3.273                       \\
               & AT4      & \textcolor{red}{\ding{55}} & \textcolor{red}{\ding{55}} & 13.773                       & \textbf{51.139}              & 15.025                       & \textbf{36.538}              & 6.231                       \\
        % \hdashline[0.5pt/1.5pt]                                                                                                                                                                                                                                       \\[-2ex]
               & AT5      & \textcolor{black}{SVM}     & \checkmark                 & \textbf{3.178}               & 66.162                       & 3.791                        & 49.667                       & \textbf{2.505}              \\
        \midrule
        \multirow{3}{*}{Engine}
        % Variance representation of epistemic
               & E1       & \checkmark                 & \checkmark                 & \textbf{0.407}               & 67.348                       & \textbf{0.501}               & 38.358                       & \textbf{0.656}              \\
               & E2       & \checkmark                 & \textcolor{red}{\ding{55}} & 2.144                        & 65.380                       & 2.258                        & 37.041                       & 1.590                       \\
               & E3       & \textcolor{red}{\ding{55}} & \checkmark                 & 0.731                        & 57.972                       & 0.509                        & 35.885                       & 0.803                       \\
               & E4       & \textcolor{red}{\ding{55}} & \textcolor{red}{\ding{55}} & 4.788                        & \textbf{54.298}              & 2.531                        & \textbf{34.632}              & 2.925                       \\
        % \hdashline[0.5pt/1.5pt]                                                                                                                                                                                                                                       \\[-2ex]
               & E5       & \textcolor{black}{SVM}     & \checkmark                 & 0.725                        & 64.285                       & 0.509                        & 41.701                       & 0.950                       \\
        \bottomrule
    \end{tabular}

\end{table*}

\section{Discussion}
\label{sec:Discussion}

Several key assumptions underpin the logic of ensemble probabilistic neural networks in this work. 
One of the main statistical assumptions is that the prediction distributions follow a normal 
Gaussian distribution. Another crucial assumption is that the individual model predictions 
are independent, however, this assumption is not strictly valid, 
as all models in the ensemble are trained on the same dataset. 
Despite this limitation, ensemble method remains effective by utilizing the randomness 
introduced through different weight initialization and training dynamics.

The performance of the adaptive thresholding mechanism within the proposed framework is 
highly dependent on the availability of correlation to aleatoric uncertainty 
source as a feature in the set of 
available model inputs. If the necessary excitation required for identifying variance in 
predictions is absent from the input features, the model will struggle to learn 
meaningful variance estimates.

While the use of an OOD indicator helps to reduce false alarms by filtering 
anomalies arising from operational shifts, it can also lead to an increase in missed detections. 
This trade-off is due to the presence of two distinct types of anomalies, those caused by 
operating point changes and those stemming from actual faults. Rejecting anomalies related to 
operational shifts is beneficial for reducing false alarm rates. However, if the OOD mechanism 
also suppresses fault-induced anomalies, it may impair the diagnostic model's 
ability to detect real faults ($p_{\text{D}}$). 
In the case studies presented here, the probability of detection and hence
the overall diagnostic performance improved across all 
applications, as the number of benign anomalies outweighed the missed fault-related anomalies. 
Nevertheless, this balance is data-dependent, and in principle, the use of OOD indicators 
could degrade detection performance if critical fault signatures are mistakenly treated as outliers.
Note that structural analysis can be helpful in this context. 
While the MSO sets are derived from the structural analysis of the model, 
the selection of causality used in residual design remains a choice, 
as it influences the valid input range associated with different causalities. 
This is an important aspect that can be used to improve the 
diagnostic quality derived from data-driven residuals, as discussed in \cite{JUNG20232903}.
% \textcolor{red}{Can be backed with an illustration}

A neural network-based residual, trained on fault-free data and aligned with structural analysis causality, 
offers a powerful tool for modeling complex system behavior and 
strong candidate for residual generation of consistency based diagnosis.
However, these networks do not necessarily 
conform to the sensitivity structure defined in the decision matrix. 
Neural networks primarily learn patterns based on statistical 
correlations rather than explicitly capturing the true 
causal relationships between signals. 
When strong correlations exist between specific inputs and the target signal, 
the network may prioritize these over the underlying causal dynamics, 
potentially leading to missed detections due to the loss of 
critical connections necessary for identifying faults. 
Moreover, fault isolation in structural models typically 
relies on the assumption that the residual model accurately 
reflects the true dynamics between signals. 
However, neural networks, which often focus on correlations 
rather than causation, may violate this assumption, 
resulting in false alarms as normal variations are erroneously 
classified as faults. 
One way to address this issue, as explored by \cite{MOHAMMADI2025106283}, 
is through the use of augmented training data tailored to each residual generator. 
One benefit of employing structural analysis is the ability
to divide a system into submodels. This allows for the individual
definition of each submodel's nominal performance.
Consequently, data collected from a specific fault scenario
can serve as the nominal performance for one submodel
while representing a fault for another.
By enhancing the training set with data from a broader range of nominal operating conditions, 
including data from decoupled faults that are non-sensitive to a specific residual, 
the model can learn a more accurate representation of valid behaviors.

\section{Conclusion}
\label{sec:Conclusion}

In this paper, we have presented an uncertainty aware data-driven diagnostic framework 
for consistency based diagnosis, using ensemble probabilistic neural networks based on 
physical causal relationships. 
The uncertainty characterization is integrated into the diagnostic decision-making process, 
providing an OOD rejection mechanism that issues warnings, 
and an adaptive thresholding strategy that adjusts the diagnostic scrutiny according to the 
stochastic nature of the data.
The framework was evaluated using consistency-based performance metrics, 
and experimental results on both simulated and real-world datasets show the 
capabilities of the proposed architecture and training pipeline.

{\appendices
\section{Derivation of ensemble predictive mean and variance}
\label{app:ensemble_moments}

We consider an ensemble of $M$ independent probabilistic models, each producing a Gaussian predictive distribution $\mathcal{N}(\hat{\mu}_i, \hat{\sigma}_i^2)$ for a given input. Treating this ensemble as a uniformly weighted mixture of Gaussians, we approximate the aggregate predictive distribution by matching its first and second moments.
The predictive mean of the mixture is obtained by the linearity of expectation:
\[
    \hat{\mu}_{\ast} = \mathbb{E}[Y] = \frac{1}{M} \sum_{i=1}^{M}  \mathbb{E}[Y \mid i] =\frac{1}{M} \sum_{i=1}^{M} \hat{\mu}_i.
\]
To compute the predictive variance, we apply the law of total variance:
\[
    \hat{\sigma}_{\ast}^2 = \mathrm{Var}(Y) = \mathbb{E}[\mathrm{Var}(Y \mid i)] + \mathrm{Var}(\mathbb{E}[Y \mid i]).
\]
The first term, $\mathbb{E}[\mathrm{Var}(Y \mid i)]$, corresponds to the average of the 
individual predictive variances.
Using the definition of variance ($\mathrm{Var}(X) = \mathbb{E}\left[(X - \mathbb{E}[X])^2\right]$, 
and substituting $X = \hat{\mu}_i$)
the second term becomes:
\[
    \mathrm{Var}(\mathbb{E}[Y \mid i]) = \mathbb{E} \left[ \left( \hat{\mu}_i - \hat{\mu}_{\ast} \right)^2 \right] = \frac{1}{M} \sum_{i=1}^{M} \left( \hat{\mu}_i - \hat{\mu}_{\ast} \right)^2.
\]
Putting both terms together, the ensemble predictive variance becomes:
\[
    \hat{\sigma}_{\ast}^2 = \frac{1}{M} \sum_{i=1}^{M} \hat{\sigma}_i^2 + \frac{1}{M} \sum_{i=1}^{M} (\hat{\mu}_i - \hat{\mu}_{\ast})^2.
\]
\section{Two tank}
\label{app:two_tank}

The water tank system used in this study is illustrated in Figure~\ref{fig:two_tank_schematic}.
The system consists of two vertically aligned tanks, with a pump delivering
water into the upper tank (Tank 1). The water then flows to the lower tank (Tank 2),
and exits from there. The control input $u$ regulates the pump based on the level in Tank 1.
The system includes four sensor measurements: $y_1$ measures the water
level in Tank 1, $y_2$ measures the water level in Tank 2, $y_3$ captures
the flow between Tank 1 and Tank 2, and $y_4$ monitors the outflow from Tank 2.
Table~\ref{tab:two_tank_faults} summarizes the types of faults introduced in the simulation. These cover sensor faults, actuator faults, leakages, and flow obstructions.
The governing equations used to create the structural model for this case study can be found in \cite{MOHAMMADI20232909}.
The selection of residuals is based on the maximum isolability and detectability of faults.

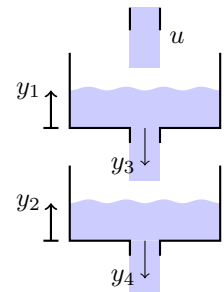
\begin{figure}
    \centering
    \begin{tikzpicture}
        
        \path[fill=blue,opacity=0.2] (-0.2,1.8+\tap) -- (-0.2,1.7+\tap) -- (0.2,1.7+\tap) -- (0.2,1.8+\tap) -- cycle;

        % \path[fill=blue, opacity=0.2, path fading = south] (-0.2,1.7+\tap) -- (-0.2,1.0+\tap) -- (0.2,1.0+\tap) -- (0.2,1.7+\tap) -- cycle;
        \path[fill=blue, opacity=0.2] (-0.2,1.7+\tap) -- (-0.2,1.0+\tap) -- (0.2,1.0+\tap) -- (0.2,1.7+\tap) -- cycle;

        \draw[thick] (-0.2,1.8+\tap) -- (-0.2,1.5+\tap);

        \draw[thick] (0.2,1.8+\tap) -- (0.2,1.5+\tap);

        \draw (0.2,1.4+\tap)  node[right]{$u$};

        % Water level

        \path[fill=blue,opacity=0.2] (-1,0) -- (-1,0.5) to[out=0, in=180] (-0.75, 0.55) to[out=0, in=180] (-0.5, 0.50) to[out=0, in=180] (-0.25, 0.55) to[out=0, in=180] (0, 0.50)

        to[out=0, in=180] (0.25, 0.55) to[out=0, in=180] (0.50, 0.50) to[out=0, in=180] (0.75, 0.55) to[out=0, in=180] (1, 0.50) -- (1,0) -- cycle;

        % \path[fill=blue,opacity=0.2,path fading = south] (-0.2,0) -- (-0.2,-0.7) -- (0.2, -0.7) -- (0.2, 0) -- cycle;
        \path[fill=blue,opacity=0.2] (-0.2,0) -- (-0.2,-0.7) -- (0.2, -0.7) -- (0.2, 0) -- cycle;

        % \draw (0.2,-0.35)  node[right]{$f_{c1}$};
        
        \draw[->] (0, 0) -- (0, -0.5) node[left]{$y_3$};

           % Tank 1
    
        \draw[thick] (-1,1) -- (-1,0) -- (-0.2,0) -- (-0.2,0) -- (-0.2, -0.2);
    
        \draw[thick] (1,1) -- (1,0) -- (0.2,0) -- (0.2,0) -- (0.2, -0.2);
    
        % \draw[thick] (1.4, 0) -- (1.6, 0);
    
        % \draw[thick, ->] (1.5, 0) -- (1.5, 0.5) node[right]{$h_1$};
        
        \draw[thick] (-1.15, 0) -- (-1.35, 0);
    
        \draw[thick, ->] (-1.25, 0) -- (-1.25, 0.5) node[left]{$y_1$};    

        \draw[thick] (-1.15, -1.5) -- (-1.35, -1.5);

        \draw[thick, ->] (-1.25, -1.5) -- (-1.25, -1) node[left]{$y_2$};    
    
        % Water level
    
        \path[fill=blue,opacity=0.2] (-1,0+\tanktwo) -- (-1,0.5+\tanktwo) to[out=0, in=180] (-0.75, 0.55+\tanktwo) to[out=0, in=180] (-0.5, 0.50+\tanktwo) to[out=0, in=180] (-0.25, 0.55+\tanktwo)
    
        to[out=0, in=180] (0, 0.50+\tanktwo) to[out=0, in=180] (0.25, 0.55+\tanktwo) to[out=0, in=180] (0.50, 0.50+\tanktwo) to[out=0, in=180] (0.75, 0.55+\tanktwo)

        to[out=0, in=180] (1, 0.50+\tanktwo) -- (1,0+\tanktwo) -- cycle;

        % \path[fill=blue,opacity=0.2,path fading = south] (-0.2,0+\tanktwo) -- (-0.2,-0.7+\tanktwo) -- (0.2, -0.7+\tanktwo) -- (0.2, 0+\tanktwo) -- cycle;         
        \path[fill=blue,opacity=0.2] (-0.2,0+\tanktwo) -- (-0.2,-0.7+\tanktwo) -- (0.2, -0.7+\tanktwo) -- (0.2, 0+\tanktwo) -- cycle;

        % \draw (0.2,-0.35+\tanktwo)  node[right]{$f_{c2}$}; 
        
        \draw[thin, ->] (0, \tanktwo) -- (0, -0.5+\tanktwo) node[left]{$y_4$};    
    
        % Tank 2
        \draw[thick] (-1,1+\tanktwo) -- (-1,0+\tanktwo) -- (-0.2,0+\tanktwo) -- (-0.2,0+\tanktwo) -- (-0.2, -0.2+\tanktwo);
        \draw[thick] (1,1+\tanktwo) -- (1,0+\tanktwo) -- (0.2,0+\tanktwo) -- (0.2,0+\tanktwo) -- (0.2, -0.2+\tanktwo);
        % \draw[thick] (1.4, 0+\tanktwo) -- (1.6, 0+\tanktwo);
        % \draw[thick, ->] (1.5, 0+\tanktwo) -- (1.5, 0.5+\tanktwo) node[right]{$h_2$};         
\end{tikzpicture}
    \caption{Schematic view of two coupled water tanks.}
    \label{fig:two_tank_schematic}
\end{figure}

\begin{table}
    \centering
    \caption{Possible faults in the two-tank system}
    \begin{tabular}{ll}
        \hline
        \textbf{Fault ID} & \textbf{Description}                          \\
        \hline
        Fa                & Actuator fault in the pump                    \\
        Fh1               & Fault in water level sensor $y_1$ (Tank 1)    \\
        Fh2               & Fault in water level sensor $y_2$ (Tank 2)    \\
        Ff1               & Fault in flow sensor $y_3$ (between tanks)    \\
        Ff2               & Fault in flow sensor $y_4$ (outflow)          \\
        Fl1               & Leakage between Tank 1 and sensor 3           \\
        Fl2               & Leakage between sensor 3 and Tank 2           \\
        Fl3               & Leakage between Tank 2 and sensor 4           \\
        Fc1               & Partial obstruction between Tank 1 and Tank 2 \\
        Fc2               & Partial obstruction after Tank 2              \\
        \hline
    \end{tabular}
    \label{tab:two_tank_faults}
\end{table}

\section{Aftertreatment system}
\label{app:AT}

The aftertreatment system case study focuses on the Selective Catalytic Reduction (SCR) subsystem in a diesel engine, which reduces nitrogen oxide (NOx) emissions to comply with regulatory standards. A reducing agent, typically urea, is injected into the exhaust gas stream where it reacts with NOx to produce nitrogen and water.
Figure~\ref{fig:AT_schematic} illustrates the dosing subsystem responsible for controlling the urea injection. Urea is pumped from a storage tank through a series of hoses and filters to a dosing chamber, where it is injected into the exhaust via a nozzle.
The governing equations used to create the structural model for this case study can be found in \cite{MOHAMMADI2025106283}.

The system features several key signals: three pressure sensors positioned before the pump filter ($y_{p,tp}$), after the pump ($y_{p,ap}$), and inside the dosing unit ($y_{p,du}$), as well as two signals from the ECU, the pump speed ($n_p$) and the dosing control signal ($u_{DC}$). The control signal is pulse-width modulated, and $u_{DC}$ refers to its duty cycle.

\begin{figure}[h]
    \centering
    \includegraphics[width=0.9\linewidth]{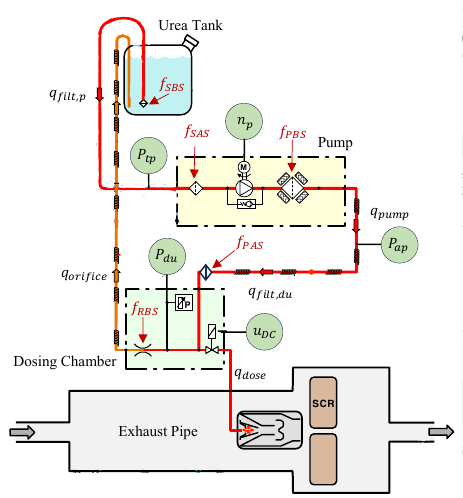}
    \caption{Schematic view of components in the aftertreatment system model. The red line represents the supply pipe, and the orange line represents the return pipe. Available signals are shown in green, and fault locations are indicated in red.}
    \label{fig:AT_schematic}
\end{figure}

Operational data was collected at a sampling rate of 10\,Hz under different drive cycles simulating both nominal and faulty conditions. Due to hardware and bandwidth constraints, signals were initially recorded at varying frequencies, which were later synchronized via interpolation.
Nominal operation includes a baseline drive cycle followed by a pressure-altering cycle designed to excite dynamic system responses.
Clogging faults were identified by the industrial partner as the most critical for system maintenance. Consequently, data from a range of clogging scenarios was collected by physically replacing standard components with partially obstructed ones. These faults were categorized by location and severity: suction-side clogging (SAS, SBS), return-side clogging (RBS), and pressure-side clogging (PAS, PBS). Some scenarios include multiple severity levels (mild, moderate, and severe), as summarized in Table~\ref{tb:faults_appendix}.

\begin{table*}
    \centering
    \caption{Summary of fault scenarios and component failures in aftertreatment system.}
    \label{tb:faults_appendix}
    \begin{tabular}{c c c c c}
        \toprule
        Label & Symbol & Intensity & Description & Component \\
        \midrule
        % NF & -- & -- & Nominal behavior & -- \\
        PAS & $f_{PAS}$ & Mild & Clogging after pressure sensor & Heating circuit \\
        PBS & $f_{PBS}$ & Mild & Clogging before pressure sensor & Pump main filter \\
        RBS1/2/3 & $f_{RBS}$ & Mild/Mod/Severe & Return pipe clogging & Orifice \\
        SAS1/2/3 & $f_{SAS}$ & Mild/Mod/Severe & Suction pipe clogging after sensor & Pump input filter \\
        SBS1/2/3 & $f_{SBS}$ & Mild/Mod/Severe & Suction pipe clogging before sensor & Urea tank filter \\
        \bottomrule
    \end{tabular}
\end{table*}

The selection of residuals is based on the maximum isolability and detectability of faults,
following the approach inspired by \cite{MOHAMMADI2025106283}.
The illustration of sensitivity and diagnosis results for the aftertreatment system
is shown in Figure~\ref{fig:AT_sensitivity} and Figure~\ref{fig:AT_isolation_performance}, respectively.

\begin{figure}
    \begin{center}
    \includegraphics[width=\columnwidth]{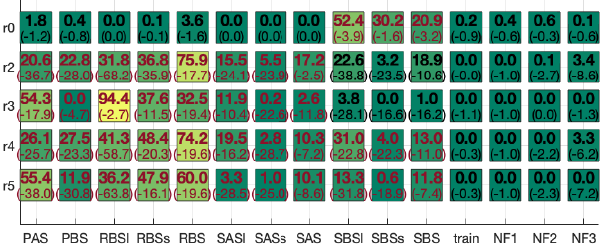} 
    \caption{Aftertreatment systems sensitivity matrix.} 
    \label{fig:AT_sensitivity}
    \end{center}
\end{figure}

\begin{figure}
    \begin{center}
    \includegraphics[width=\columnwidth]{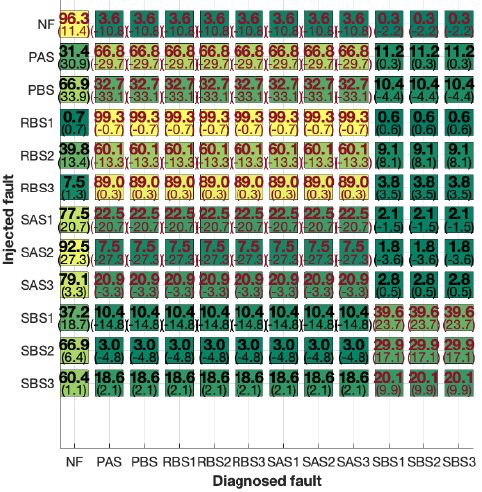} 
    \caption{Aftertreatment systems fault isolation performance matrix.} 
    \label{fig:AT_isolation_performance}
    \end{center}
\end{figure}
\section{Automotive gasoline engine air-path}
\label{app:engine_airpath}

This case study is based on the LiU-ICE Industrial Fault Diagnosis Benchmark \cite{jung2024benchmark},
focusing on the air path of a turbocharged gasoline engine.
The air path is a complex subsystem involving components such as the air filter,
throttle, intercooler, intake manifold, and turbocharger.
Figure~\ref{fig:Engine_schematic} illustrates the schematic of the automotive gasoline engine air-path system.
% Accurate diagnosis of faults within this subsystem is essential for maintaining engine performance 
% and meeting emission standards. 
The governing equations used to create the structural model for this case study can be found in \cite{jung2024benchmark}.
The set of sensors and actuators, summarized in Table~\ref{tab:engine_signals}.

\begin{figure}[h]
    \centering
    \includegraphics[width=0.9\linewidth]{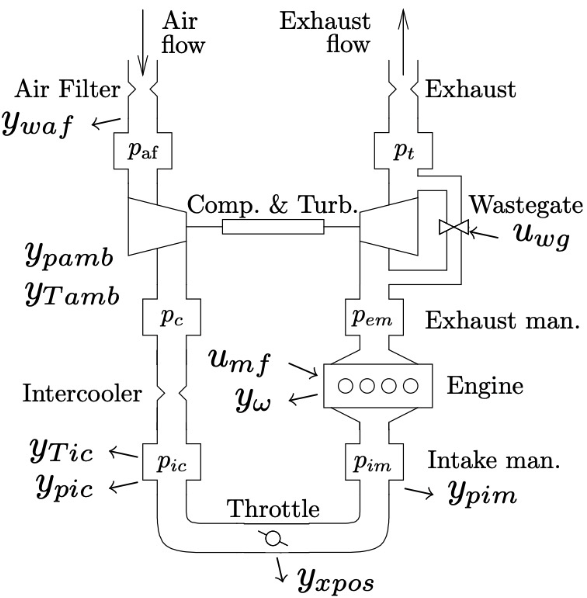}
    \caption{Schematic of the automotive gasoline engine air-path system \cite{jung2024benchmark}.}
    \label{fig:Engine_schematic}
\end{figure}

\begin{table}[h]
    \centering
    \caption{Sensors and actuators in the gasoline engine air-path system.}
    \label{tab:engine_signals}
    \begin{tabular}{ll}
        \toprule
        \textbf{Signal} & \textbf{Description}                  \\
        \midrule
        \multicolumn{2}{l}{\textit{Sensors}}                    \\
        $y_{pic}$       & Intercooler pressure                  \\
        $y_{Tic}$       & Intercooler temperature               \\
        $y_{pim}$       & Intake manifold pressure              \\
        $y_{waf}$       & Mass flow through the air filter      \\
        $y_{xpos}$      & Throttle actuator position            \\
        $y_{\omega}$    & Engine speed                          \\
        $y_{pamb}$      & Ambient pressure                      \\
        $y_{Tamb}$      & Ambient temperature                   \\
        \midrule
        \multicolumn{2}{l}{\textit{Actuators}}                  \\
        $u_{mf}$        & Requested injected fuel mass          \\
        $u_{wg}$        & Requested wastegate actuator position \\
        \bottomrule
    \end{tabular}
\end{table}

Data was collected from an engine test bench under various operating conditions, including both nominal and faulty scenarios. The datasets encompass different driving cycles, capturing a wide range of engine behaviors. Each dataset contains approximately 30 minutes of data sampled at 20\,Hz.
The benchmark includes several fault types, summarized in Table~\ref{tab:engine_faults}. Sensor faults are introduced as multiplicative deviations, while the leakage fault is physically induced.
Each fault scenario is introduced approximately 120 seconds into the dataset and persists until the end.
The evaluation data includes one to two datasets for each fault type, with varying fault magnitudes.
Due to the large number of residual candidates for this case,
a minimal set of residuals that achieves the highest detection and isolation performance is selected.
The illustration of sensitivity and diagnosis results for Engine
is illustrated in Figure~\ref{fig:Engine_sensitivity} and
Figure~\ref{fig:Engine_isolation_performance} respectively.

\begin{table}[h]
    \centering
    \caption{Fault scenarios in the gasoline engine air-path benchmark.}
    \label{tab:engine_faults}
    \begin{tabular}{ll}
        \toprule
        \textbf{Component} & \textbf{Fault location}         \\
        \midrule
        $fpim$             & Intake manifold pressure sensor \\
        $fpic$             & Intercooler pressure sensor     \\
        $fwaf$             & Air mass flow sensor            \\
        $fiml$             & Leakage in the intake manifold  \\
        \bottomrule
    \end{tabular}
\end{table}

\begin{figure}
    \begin{center}
        \includegraphics[width=\columnwidth]{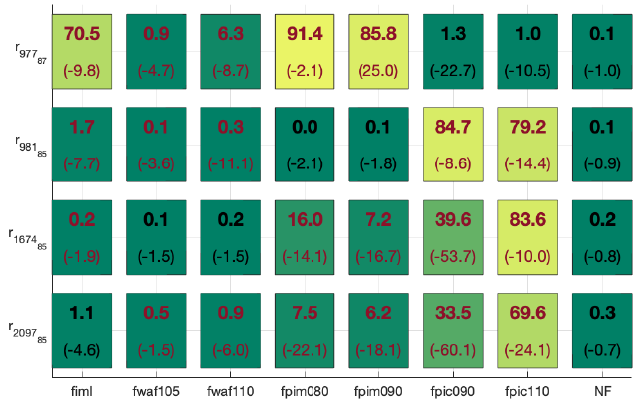}
        \caption{Engine sensitvity matrix.}
        \label{fig:Engine_sensitivity}
    \end{center}
\end{figure}

\begin{figure}
    \begin{center}
        \includegraphics[width=\columnwidth]{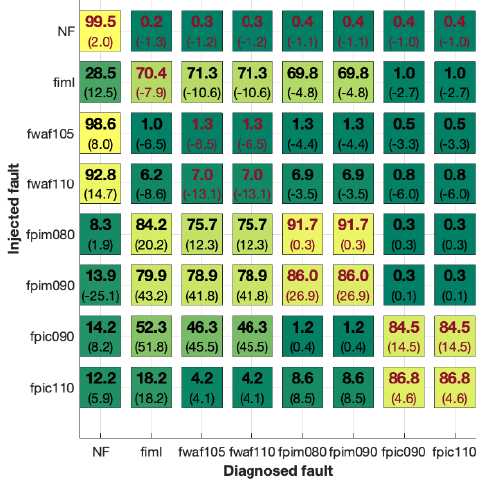}
        \caption{Engine fault isolation performance matrix}
        \label{fig:Engine_isolation_performance}
    \end{center}
\end{figure}

}

\bibliographystyle{IEEEtran}
\bibliography{myreferences}

\end{document}